\documentclass[letterpaper]{article} 
\usepackage{aaai24}  
\usepackage{times}  
\usepackage{helvet}  
\usepackage{courier}  
\usepackage[hyphens]{url}  
\usepackage{graphicx} 
\usepackage{natbib}  
\usepackage{caption} 
\usepackage{algorithm}
\usepackage{xspace}
\newcommand{\method}{\texttt{ConSequence}\xspace}

\usepackage{booktabs}       
\usepackage{amsfonts}       
\usepackage{eqnarray, amsmath}
\usepackage{algorithmicx}
\usepackage{algpseudocode}
\usepackage{xcolor}
\usepackage{caption}
\usepackage{subcaption}
\usepackage{multirow}
\usepackage{tikz}
\usepackage{array, makecell}
\usepackage{enumitem}
\usepackage{relsize}
\usepackage{amsthm}

\newtheorem{theorem}{Theorem}
\newtheorem{definition}{Definition}
\setlist[itemize]{leftmargin=*}
\setlist[enumerate]{leftmargin=*}
\setlist{nosep}

\title{ConSequence: Synthesizing Logically \underline{Con}strained \underline{Sequence}s for Electronic Health Record Generation}
\author {
    Brandon Theodorou\textsuperscript{\rm 1, 2},
    Shrusti Jain\textsuperscript{\rm 1},
    Cao Xiao\textsuperscript{\rm 3},
    Jimeng Sun\textsuperscript{\rm 1, 2}
}
\affiliations {
    \textsuperscript{\rm 1}University of Illinois at Urbana?Champaign, Urbana, IL, United States\\
    \textsuperscript{\rm 2}Medisyn Inc., Las Vegas, NV, United States\\
    \textsuperscript{\rm 3}GE Healthcare, Chicago, IL, United States\\
    bpt3@illinois.edu, jimeng@illinois.edu
}
\begin{document}

\maketitle

\begin{abstract}
Generative models can produce synthetic patient records for analytical tasks when real data is unavailable or limited. However, current methods struggle with adhering to domain-specific knowledge and removing invalid data. We present \method, an effective approach to integrating domain knowledge into sequential generative neural network outputs. Our rule-based formulation includes temporal aggregation and antecedent evaluation modules, ensured by an efficient matrix multiplication formulation, to satisfy hard and soft logical constraints across time steps. Existing constraint methods often fail to guarantee constraint satisfaction, lack the ability to handle temporal constraints, and hinder the learning and computational efficiency of the model. In contrast, our approach efficiently handles all types of constraints with guaranteed logical coherence. We demonstrate \method's effectiveness in generating electronic health records, outperforming competitors in achieving complete temporal and spatial constraint satisfaction without compromising runtime performance or generative quality. 
Specifically, \method successfully prevents all rule violations while improving the model quality in reducing its test perplexity by 5\% and incurring less than a 13\% slowdown in generation speed compared to an unconstrained model.
\end{abstract}

\section{Introduction} 
Sequential data generation applications are used in various fields, such as healthcare \cite{MedGAN, EVA, SynTEG, theodorou2023synthesize}, finance \cite{assefa2020generating, dogariu2022generation}, natural language processing \cite{gatt2018survey, dong2022survey, reiter1997building}, and computer vision \cite{tulyakov2018mocogan, yang2022diffusion, ho2022video}. In these applications, generative models demonstrate the capability to produce synthetic data closely resembling real-world datasets.
Notably, language models have showcased remarkable achievements in domains like text generation~\cite{chowdhery2022palm, thoppilan2022lamda, GPT3} and health record generation~\cite{theodorou2023synthesize}, primarily attributed to their adeptness in accurate next-token forecasting.
Beyond mimicking real data, generative models often require adhering to specific rules and temporal dependencies unique to an application domain. 
For instance, in medicine, generative models should follow logical constraints such as indications (the reason for using a specific treatment for a disease) and contraindications (the reason that makes a particular treatment inadvisable for that disease) to create realistic sequential data.

To support domain-specific applications, generated samples must not only approximate the true underlying data distribution but also encapsulate the relationships and dependencies encoded as rules derived from external knowledge. Failing to meet these constraints can hinder efficacy and deter adoption. Various current approaches attempt to ensure logical consistency in neural network outputs. These methods encompass strategies such as integrating rule constraints into the loss function \cite{xu2018semantic, fischer2019dl2}, employing post-processing modules designed to rectify model output violations  \cite{manhaeve2018deepproblog, hoernle2022multiplexnet, ahmed2022semantic}, or directly adding model components that ensure the final weights and outputs align with domain knowledge \cite{towell1994knowledge, avila1999connectionist, giunchiglia2021multi}. 
Despite the existing effort in constraint enforcement, sequential data generation still faces several challenges in aligning with real-world knowledge.
\begin{itemize}
    \item \textbf{Inadequate treatment of temporal logical constraints}. Existing models struggle with temporal constraints, which can be static or evolve across multiple time steps. These constraints are key to enhancing accuracy and reliability in sequential generation models. Yet, no existing models effectively manage these constraints as the information conditioning them may not be accessible at each time step.
    \item \textbf{Efficiency and scalability}. Sequential tasks typically encompass a multitude of time steps and entail a substantial output dimensionality, underscoring the need for a streamlined approach to facilitate large-scale sequence generation. Current methods can be slow and computationally demanding, impairing real-world generation speed~\cite{bond2021deep}. Thus, an effective solution should not only be asymptotically efficient but also maintain high generation speeds for practical use.
    \item \textbf{Difficulty in achieving full logical consistency}. Realism in sequential data generation is paramount, but logical inconsistencies can compromise it. Hence, encoding domain knowledge without error is essential. For full trustworthiness, methods must uphold all logical constraints; even one violation can erode user confidence. This is especially vital in critical areas like healthcare.
\end{itemize}

In this paper, we present a method called \method for addressing the challenges of sequential constrained knowledge infusion. \method can handle entailment formatted rules with both soft and hard logical constraints, covering static, temporal, and combined antecedents.
\begin{itemize}
    \item \textbf{Temporal antecedent compilation via attentive history aggregation}. We employ an attention-based temporal aggregator for effectively consolidating historical data at each time step, supporting both absolute and relative aggregation while accommodating sequence variations.
    \item \textbf{Deterministic constraint execution via rule neuron}. We introduce a graphical module that verifies each antecedent component, providing absolute constraint satisfaction.
    \item \textbf{Parallel constraint application via efficient GPU implementation}. We represent the mentioned components as weight matrices. These are then seamlessly incorporated as a constraint head in neural network models, processing many rules and records in parallel, resulting in minimal to no GPU slowdown and maximizing overall efficiency.
\end{itemize}

To assess its effectiveness, we undertake a series of experiments on a benchmark task employing the cutting-edge health record generation model, HALO \cite{theodorou2023synthesize}, across two real-world datasets. We benchmark our method against prevailing constraint enforcement approaches, evaluating them on: (1) rule violation count, (2) overall model quality, and (3) generative efficiency.

Our results demonstrate that \method outperforms all other methods over all three criteria. While most baselines result in over 25\% of their generated records for one dataset being invalid, \method successfully prevents any rule violations.
Furthermore, \method improves the model quality and incurs less than a 13\% slowdown in generation speed compared to an unconstrained model. 
In comparison, all of the baselines that are not loss-based exhibit slowdowns exceeding 37\%, with the majority performing even more poorly. 
This highlights the effectiveness of \method for generating high-quality health records while satisfying domain-specific constraints.

\section{Related Work}
Various strategies have been proposed to apply logical constraints to neural network models, especially in  multi-label predictions~\cite{dash2022review,giunchiglia2022deep}. One  method is to include logical constraints in the loss function during training, as regularization or penalty terms \cite{xu2018semantic, fischer2019dl2}. Another approach involves adjusting the model's weights to satisfy the constraints \cite{towell1994knowledge, avila1999connectionist, ahmed2022semantic}. Alternatively, some techniques focus on mapping the model's outputs to meet the constraints \cite{manhaeve2018deepproblog, hoernle2022multiplexnet, giunchiglia2021multi}.

Despite the range of approaches, the ultimate goal is consistent: to ensure the model's outputs align with logical constraints, enhancing interpretability and trustworthiness. While progress has been made in applying constraint methods to various applications, adapting these techniques to our specific task faces challenges. For instance, loss-based approaches \cite{xu2018semantic, fischer2019dl2} reduce constraint violations but do not guarantee logical consistency, potentially limiting practical use for our needs. Additionally, certain methods ensuring consistency \cite{giunchiglia2021multi, hoernle2022multiplexnet} can be slow, which might not suit large-scale sequential generation settings.

Moreover, our task involves integrating temporal rules that may change over time, without their logical precursors necessarily present at each step. Although some methods have applied constraints to sequential text and molecule data \cite{hokamp2017lexically, liu2020chance}, none have specifically addressed the challenge of satisfying general temporal logical constraints efficiently. This highlights the need for innovative approaches capable of handling complex temporal constraints in a computationally efficient manner.

\section{Problem Formulation}
We present the problem formulation for the task of longitudinal patient record generation. 
We first define patient data.

\begin{definition}[\textbf{Sequential Patient Health Record Data}]
A patient record is a time-sequenced series of visits, denoted as $\mathcal{P} = \mathcal{V}^{(1)}, \mathcal{V}^{(2)}, \cdots, \mathcal{V}^{(T)}$. Each visit, $\mathcal{V}^{(t)}$, consists of a variable number of unique medical codes $c^{(t)}_i$ from a set $\mathcal{C}$. These codes, $(c^{(t)}_1, c^{(t)}_2, \cdots c^{(t)}_k)$, encode medical information such as diagnoses, procedures, and medications.
\end{definition} 
To prepare for machine learning models, we transform $\mathcal{P}$ into a matrix representation 
$\mathbf{P} = [\mathbf{v}_1, \mathbf{v}_2, \cdots, \mathbf{v}_{T}]$ for a patient with $T$ visits. Each visit $\mathbf{v}_t \in \mathbb{R}^{|\mathcal{C}|}$ is a multi-hot binary vector, with $c_t^i \in \{0,1\}$ representing the presence of the $i$-th code in the $t$-th visit of $\mathcal{P}$. Static features such as gender, ethnicity, and birth year are encoded in the first visit $\mathcal{V}^{(1)}$.

In synthetic patient generation, our aim is to create synthetic patient sequences $\mathbf{P}'$ from scratch, given a real patient dataset $\mathcal{D}$ to train on, that mimic real records $\mathbf{P} \in \mathcal{D}$ and offer equivalent downstream utility. We incorporate real-world knowledge through logical constraints $\mathcal{R}$ to guide generation. These constraints, defined as a series of logical entailment statements over codes and visits, may be grounded in a single time step or span multiple time steps. We encode them in a format called Conjunctive Implicative Form.

\begin{definition}[\textbf{Conjunctive Implicative Form}]
A statement is in Conjunctive Implicative Form (CIF) if it is of the form $a_1 \land a_2 \land \ldots \land a_k \implies a_{k+1}$ for $k \geq 0$, where $a_i$ is a literal such that $a_i = x_j$ or $a_i = \lnot x_j$ for a variable $x_j$. 
\end{definition}

\noindent\textbf{Encoding Rules}. Our set of variables consists of generated codes across all timesteps, that is, all $c_t^i$. We let $a_t^i$ denote either of $c_t^i$ and $\lnot c_t^i$. Given this, our rules must be in one of the following formats.

\begin{enumerate}
    \item $True \Rightarrow a_t^j$, a rule with an empty antecedent, though this does not generally appear in practice.
    \item $a_t^i \Rightarrow a_t^j$, represents a hierarchical relationship: a specified truth value of $c_t^i$ necessitates one of $c_t^j$. For example, if a patient is taking insulin, they must have diabetes.
    \item $a_t^i \land a_t^j \land \cdots \land a_t^k \Rightarrow a_t^l$, which extends the former to relate the truth value of a boolean combination of terms to that of a singular term. For example, if a patient has heart disease and takes a statin, they must have high cholesterol.
    \item $a_{t-1}^i \Rightarrow a_t^j$, a temporal rule relating the prior timestep and current time. For example, lifelong diseases like diabetes will always continue to appear once they begin.
    \item $a_{I}^i \Rightarrow a_t^j$, a boolean combination of terms across a set of specified previous timesteps, $I$, which relate to the current time. For example, a patient without a past history of pregnancy can't have complications during childbirth.
    \item $a_{I}^i \land \cdots \land a_{I}^j \land \cdots \land a_t^k \land \cdots \land a_t^l \Rightarrow a_t^m$, relating a boolean combination of terms across both past and current timesteps to a singular boolean variable $c_t^m$. For example, a patient with past diabetes and current numbness in their feet must have diabetic peripheral neuropathy.
\end{enumerate}

Each rule $r$ consists of a temporal component $I^{(r)} = \{ t_1, t_2, \cdots, t_{|I|} \}$ which determines the indices of the past time steps referred to, the variables in the past and present which condition the output, and the output literal\footnote{We assume rules are acyclic but make no additional assumptions, and this approach can be generalized to cyclic rules with an initial transformation and elimination step, not discussed here.}. If a rule has a non-empty temporal component, we classify it as temporal, and we otherwise consider it to be static. We use entailment statements with conjunctive antecedents to represent rules, offering conciseness, efficient processing, and aligning with logical constraints in real-world settings. In contrast, other models \cite{xu2018semantic, hoernle2022multiplexnet} use logical normal forms combining variables with AND, OR, or entailment connectors, which can be cumbersome for large outputs, without additional representational benefits. Next we show that our entailment-based format equals the representational power of alternative encoding schemes.

\begin{theorem}
Given a boolean expression $C_1 \land \dots \land C_k$ in conjunctive normal form, there exists an equivalent set of statements in conjunctive implicative form.
\end{theorem}

\begin{proof}
An expression in conjunctive normal form is the conjunction of one or more clauses $C_1 \land \dots \land C_k$, where each clause $C_i$ is the disjunction of one or more literals $a_{i, 1} \lor \dots \lor a_{i, m}$. Given an expression in conjunctive normal form, we proceed as follows. First, we convert each clause $C_i = a_{i,1} \lor \dots \lor a_{i,m}$ to to the equivalent statement $\lnot (a_{i, 1} \lor \dots \lor a_{i, m-1}) \implies a_{i,m}$. Applying De Morgan's Laws, we can further convert this to the statement in conjunctive implicative form $\lnot a_{i, 1} \land \dots \land \lnot a_{i, m-1} \implies a_{i, m}$. That is, the converted expression is now the conjunction of conjunctive implicative statements. Within our format, the conjunction of statements is implicitly represented through their mutual presence within a set, so $f$ is logically equivalent to the set of statements in conjunctive implicative form  $\{(\lnot a_{1, 1} \land \dots \land \lnot a_{1, m-1} \implies a_{1, m}), \dots, (\lnot a_{k, 1} \land \dots \land \lnot a_{k, p-1} \implies a_{k, p})\}$ formed through this conversion.
\end{proof}

\section{\method Method}

\begin{figure*}
    \centering
    \includegraphics[width=0.89\textwidth]{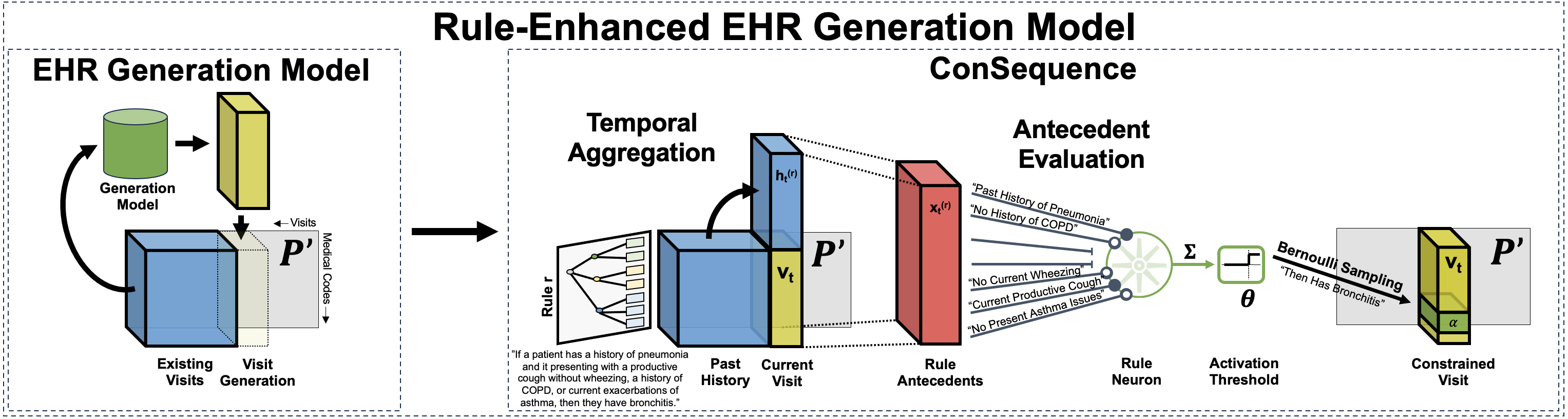}
    \caption{\method refines EHR generation models by incorporating rule-based prior knowledge. The rule evaluation process aggregates past history and feeds it with the current visit to the rule neuron as the logical antecedent. If the neuron's activation threshold is met, it fires and overrides the prior model's output to ensure logical consistency in the generated EHRs}
    \label{fig:ConSequence}
\end{figure*}

We introduce our \method method (seen in Figure \ref{fig:ConSequence}) for generating logically constrained sequential data. \method is designed to directly mirror the logical entailment process. It involves two modules: temporal aggregation and antecedent evaluation, followed by a matrix multiplication-based approach for efficient parallel constraint application optimized for modern GPU architectures.

\subsection{A. Temporal Aggregation}
To support temporal rules, we first conduct temporal sequence aggregation. Specifically, given the sequence of all visits $\mathbf{P} = [\mathbf{v}_1, \mathbf{v}_2, \dots, \mathbf{v}_T]$, where $\mathbf{v}_t$ is the visit at time $t$, we consider a temporal rule $r \in \mathcal{R}$ that contains a temporal component $I^{(r)}$. $I^{(r)}$ refers to the set of indices of past time steps whose visits are referenced within $r$, and may be both absolutely (e.g. the second time step) and relatively (e.g. the previous time step) numbered, where we represent relative numberings as negative indices. We define $f_{I^{(r)}}$ as a function that maps a time step $t$ to a set of referenced past time steps using $I^{(r)}$, converting relatively numbered visits to absolutely numbered ones and handling the inability of time steps to reference themselves and future visits as a part of their past history. For example, if $I^{(r)} = \{ -1, 1, 4 \}$, then we have $
f_{I^{(r)}}(1) = \{\},    f_{I^{(r)}}(2) = \{ 1 \},  f_{I^{(r)}}(3) = \{ 1, 2 \},  f_{I^{(r)}}(10) = \{ 1, 4, 9 \} $.
Next, we aggregate the relevant past visits according to $f_{I^{(r)}}(t)$. We define binary mask vector $\mathbf{m}_t^{(r)}$ which varies across both rules and time steps:
\begin{equation}
    m_{t,i}^{(r)} =
    \begin{cases}
    1 & \text{if } i \in f_{I^{(r)}}(t) \\
    0 & \text{otherwise}
    \end{cases}
    \quad\text{where } \mathbf{m}_t^{(r)} \in \{0,1\}^{T}.
\end{equation}

We then construct the aggregated history representation $\mathbf{h}_t^{(r)} \in \mathbb{R}^{|\mathcal{C}|}$ for time step $t$ and rule $r$ by:

\begin{equation}
\mathbf{h}_t^{(r)} = {\bigcup}_{i=1}^{t-1} m_{t,i}^{(r)} \mathbf{v}_i
\end{equation}

Here, $\bigcup$ denotes the boolean OR operation, and $\mathbf{h}_t^{(r)}$ is a static vector that aggregates the relevant codes within past visits according to the temporal component $I^{(r)}$.
This allows us to aggregate past sequences into fixed representations for each time step and rule, which are treated as standard variables for the entailment calculation in the next module.

\subsection{B. Antecedent Evaluation}
Once we have $\mathbf{h}_t^{(r)}$, we turn the problem into a standard entailment constraint problem. We solve this problem using a recurrent neuron model to evaluate the antecedent of a logical entailment and set the consequent. Each logical rule is constructed as a neuron, inspired by activation thresholds in biological and machine learning processes \cite{lin2017artificial}, that only activates when all antecedent literals are satisfied, feeding back into the input to set the consequent.

\subsubsection{Construction of the Rule Neuron} Let the variables that the rule operates over be represented by the binary vector $\mathbf{x}_t^{(r)} = \text{concat}(\mathbf{h}_t^{(r)}, \mathbf{v}_t)$ which consists of both the aggregated historical variables and current visit representation. The antecedent of one of our logical entailment rules is defined as a conjunction of literals, each of the form $x_{t,i}^{(r)}$ or $\neg x_{t,i}^{(r)}$, referring to $i$-th element of $\mathbf{x}_t^{(r)}$. Let $L = \{ x_{t,p_1}^{(r)}, \cdots, x_{t,p_p}^{(r)}, \neg x_{t,n_1}^{(r)}, \cdots, \neg x_{t,n_n}^{(r)} \}$ be the set of such literals in the antecedent which refer collectively to a subset of $\mathbf{x}_t^{(r)}$. The rule neuron accepts inputs from the variables in $\mathbf{x}_t^{(r)}$ and is constructed as follows: we denote the weight of the connection between the neuron and the $i$-th variable in $\mathbf{x}_t^{(r)}$ as $w_i^{(r)}$ and set it to $1$ if the variable is in $L$ and $-1$ if its negation is in $L$. For all other variables, the weight is set to 0. This construction is based on the logical interpretation of the antecedent, where the presence of a variable in the antecedent implies its positive contribution to the truth value, and the presence of its negation implies its negative contribution. More explicitly, we set the neuron's weights by:
\begin{equation}
    w_i^{(r)} = \begin{cases} 1 & \text{if } x_{t,i} \in L \\ -1 & \text{if } \neg x_{t,i} \in L \\ 0 & \text{otherwise} \end{cases}\\
\end{equation}

For instance, consider a rule stipulating that patients with past history of gastric ulcer and current symptom of melena, without past colorectal cancer, current hematemesis, or current esophageal varices must have gastrointestinal bleeding. The weights corresponding to the variables for past gastric ulcer and current melena in $\mathbf{x}_t^{(r)}$ would be 1 while the variables for past colorectal cancer, current hematesis, and current esophageal varices to -1. All others would be 0.

\subsubsection{Activation Process} Given the constructed neuron, we use it as follows. We first set the activation threshold $\theta^{(r)}$ to the number of positive, non-negated variables (of the form $x_{t,i}$ rather than $\neg x_{t,i}$) in $L$.
Given $\mathbf{x}_t^{(r)}$, the activation process begins. The weighted sum of the inputs is calculated by Eq.~\eqref{eq:activation}:
\begin{equation}
    s_t^{(r)} = \sum_{i=1}^{2\cdot|C|} w_i^{(r)} \cdot x_{t,i} \label{eq:activation}
\end{equation}
If the $s_t^{(r)}$ is greater than or equal to $\theta^{(r)}$, the neuron fires and sets the output variable $o^{(r)}$ in $\mathbf{v}_t$ to the output value, $\alpha^{(r)}$. Otherwise, it remains inactive.
It's important to note that, based on the construction, $\theta^{(r)}$ is reached only if all of the positive variables and none of the negations are found in the input. 
For instance, using the previously mentioned example rule, $\theta^{(r)}$ would be set to 2. $s_t^{(r)}$ would then equal 2 only if the variables for past gastric ulcer and current melena are both 1 in $\mathbf{x}_t^{(r)}$, as they are the sole potential positive contributions during the activation sum. Furthermore, the variables for past colorectal cancer, current hematesis, and current esophageal varices must all be 0 to avoid any reduction below 2 due to their corresponding weight of -1.

\subsubsection{Handling Soft Constraints} We also want to support settings where $r$ is a soft constraint with $\alpha^{(r)}$ between 0 and 1. This flexibility is essential when the relation between variables is probabilistic. For example, most cases of COPD involve a past history of smoking, but exceptions do exist. So, we might include a rule stating that a patient without a past history of smoking has just a 1\% chance of presently showing COPD and set $\alpha^{(r)} = 0.01$ accordingly. To address such cases, we add a final step before setting the output variable $o^{(r)}$ in $\mathbf{v}_t$. During training, we set the variable to the correct probability, but the output variables must equal 0 or 1 during generation. To achieve this, we sample from a Bernoulli distribution with probability $p = \alpha^{(r)}$ before setting the value, resulting in a binary output while satisfying soft constraints by aligning that output with the proper underlying probabilities. 
Note that we can also apply this sampling process to hard constraints because sampling from $p=0$ and $p=1$ always yield 0 and 1, respectively. Therefore, we can integrate this sampling process into our core activation process without special handling for different constraints types.

\subsubsection{Rule Evaluation} The rule evaluation process involves encoding the temporal history of each binary variable and feeding them, along with the current variables, to the rule neuron. The rule neuron evaluates the antecedent by computing the weighted sum of its inputs, comparing it to the activation threshold, and activating the neuron if the threshold is reached. If the neuron fires, it overrides any modeling by the preceding architecture and sets the output variable to the mandated output value, sampling from that value if during generation as opposed to training, as in Eq.~\eqref{eq:evaluation}:
\begin{equation}
    v_{t,o^{(r)}} = \begin{cases} 
        \text{Bernoulli}(\alpha^{(r)}) & \text{if } s_t^{(r)} \geq \theta^{(r)}\\
        v_{t,o^{(r)}} & \text{otherwise}
    \end{cases}\\
\label{eq:evaluation}
\end{equation}
This end-to-end process can be seen in Figure \ref{fig:ConSequence}.

\subsection{C. Parallel Constraint Application}
In the previous sections, we have presented \method in the context of a single rule and, in most cases, a single time step. During generation, we will only be concerned with the most recent visit at any given step, but during training it will be valuable to constrain all time steps simultaneously. Furthermore, applying multiple rules simultaneously is valuable for improving efficiency in any setting. So, we present a matrix multiplication formulation of \method that leverages modern GPU computing architectures to simultaneously and efficiently apply groups of rules across time steps and across batches of patient records.

\subsubsection{Rule Grouping and Combination}
We begin by organizing rules into distinct groups, denoted as $G = \{g_1, \cdots, g_n\}$ where each group $g_i = \{r_1, \cdots, r_{|g_i|}\}$ contains a set of rules that can be processed simultaneously. While the temporal aggregation module prevents the combination of rules with different temporal components, we note that most rules use one of a few common components such as the first visit, the previous visit, or all past history. With this in mind, we categorize rules according to their temporal components.

We can then combine rules in each category of temporal components into rule groups if they have different output variables and their antecedent variables do not include output variables from previous rules in the same category. To do this, we iterate through the rules of each category, adding them to the current group and starting a new group each time we encounter a rule whose input or output variables overlaps with the output of a preceding rule in the current group.

\subsubsection{Temporal Aggregation} To perform temporal aggregation across a group of rules, $g$, with a shared temporal component, we use a technique inspired by masked self-attention. It involves constructing a temporal mask matrix $\mathbf{M}^{(g)} \in \mathbb{R}^{T \times T}$, where each row $\mathbf{m}_t^{(g)} \in \mathbb{R}^T$ is a binary vector that determines which other time steps the $t$-th time step can attend to. Specifically, each $m_{t,i}^{(g)}$ in $\mathbf{m}_t^{(g)}$ is set by Eq.~\eqref{eq:matrixattend}:
\begin{equation}\label{eq:matrixattend}
    m_{t,i}^{(g)} =
    \begin{cases}
    1 & \text{if } i \in f_{I^{(g)}}(t) \\
    0 & \text{otherwise}
    \end{cases}
    \quad\text{where } \mathbf{m}_t^{(g)} \in \{0,1\}^{T}.
\end{equation}
By employing boolean matrix multiplication, we compute $\mathbf{H}^{(g)} \in \mathbb{R}^{T \times |\mathcal{C}|}$ using the patient record $\mathbf{P}$ by $\mathbf{H}^{(g)} = \mathbf{M}^{(g)}\mathbf{P}$, where each row is the aggregated temporal history vector $\mathbf{h}^{(r)}_t$ introduced previously (for any $r \in g$).

\subsubsection{Rule Execution and Output} This aggregated temporal history matrix can then be concatenated along the code dimension to $\mathbf{P}$ to form $\mathbf{X}^{(g)} = \left[ \mathbf{H}^{(g)}, \mathbf{P} \right]$. We then construct matrix $\mathbf{W}^{(g)} \in \mathbb{R}^{2\cdot|C| \times |C|}$ to represent the rule neuron weights. For each rule in the group, we set the values in the column corresponding to the output code based on the construction of $\mathbf{w}^{(r)}$ detailed earlier. 
Note that since there is no overlap in output variables within a group, each column is constructed without interference between different rules.

Utilizing matrix multiplication, we compute the antecedent sum output $\mathbf{S}^{(g)} = \mathbf{X}^{(g)} \mathbf{W}^{(g)}$, resulting in a matrix $\mathbf{S}^{(g)} \in \mathbb{R}^{T \times |C|}$ where each value whose row corresponds to time step $t$ and whose column corresponds to the output variable in rule $r$ is equal to the sum $s_t^{(r)}$ from earlier. This value is a count of how many positive literals in the corresponding antecedent were satisfied at that time step minus how many negatives were failed (0 if there was no rule corresponding to that variable). 
When this count is equal to the number of positive literals, $\theta$, we set the output code to the output value, $\alpha$. As we have multiple $\theta$ and $\alpha$ values per group, we combine the constants into vectors $\boldsymbol{\theta} \in \mathbb{R}^{|C|}$ (setting values corresponding to output codes which are not used in the rule group to -1 so that the equality is never satisfied) and $\boldsymbol{\alpha} \in \mathbb{R}^{|C|}$ respectively which may perform comparison and instantiations element-wise in parallel. To support both hard and soft constraints simultaneously, we can also first sample from the Bernoulli distribution defined by $p = \boldsymbol{\alpha}$ and set the output code to the resulting binary values. As a single step equation, this process can be summarized:
\begin{equation} \label{eq:RuleApplication}
    \mathbf{P}[\mathbf{X}^{(g)} \mathbf{W}^{(g)} == \boldsymbol{\theta}] = \text{Bernoulli}(\boldsymbol{\alpha})
\end{equation}
where the group of rules are applied simultaneously over all time steps and all records in a batch, effectively enabling parallel enforcement of rules.

\begin{table}
    \small
    \centering
    \begin{tabular}{lcc} \toprule
         & Outpatient  & Inpatient  \\\midrule
        \# Records & 1,006,321 & 46,520 \\
        Mean Visits/Record & 35.40 & 1.27 \\
        Mean Codes /Visit & 1.69 & 13.61 \\
        \# Phecodes & 1,817 & 1,610 \\
        Demographic  & 7 & 6 \\\bottomrule
    \end{tabular}
    \caption{Dataset Statistics}
    \label{tab:DataStats}
\end{table}

\begin{table*}
    \centering
    \small
    \begin{tabular}{lcccccc}
    \toprule
    & \multicolumn{3}{c}{Violations on Outpatient Dataset} & \multicolumn{3}{c}{Violations on Inpatient Dataset} \\
    & Static  & Temporal & \% Valid & Static  & Temporal   & \%  Valid       \\ \midrule
    Vanilla HALO 
    & 1265.4 (18.1)    & 17.5 (2.9)     & 87.2\% (0.002)  & 4320.6 (22.2) & 2046.2 (35.6) & 65.4\% (0.002) \\\hline
    Post Processing           & \textbf{0.0}    & \textbf{0.0}  & \textbf{100\%}   & \textbf{0.0}   & \textbf{0.0}   & \textbf{100.0\%}\\
    Semantic Loss 
    & 955.5 (9.2)    & 18.7 (3.6)       & 90.3\% (0.001)   & 3827.0 (23.3) & 1817.2 (28.8) & 65.3\% (0.002) \\
    CCN 
    & 1013.6 (15.0)    & 17.7 (2.1)     & 91.0\% (0.001)   & 3070.24 (18.3) & 2149.76 (32.5) & 70.7\% (0.002) \\
    MultiPlexNet 
    & \textbf{0.0}       & 2714.2 (3.6)    & 89.7\% (0.001)   & \textbf{0.0}     & 3304.8 (20.1)       & 72.7\% (0.002) \\
    SPL 
    & \textbf{0.0}     & 41    & 99.9\%       & \textbf{0.0}     & 11.5 (1.4)   & 99.9\% (0.0001) \\\hline
    \method           & \textbf{0.0}    & \textbf{0.0}  & \textbf{100\%}   & \textbf{0.0}    & \textbf{0.0}  & \textbf{100\%}\\\bottomrule
    \end{tabular}
    \caption{Rule Violation Counts}
    \label{tab:Violations}
\end{table*}

\subsection{D. Runtime Analysis}
\method can process, evaluate, and adhere to rules in $O(T^2|C|)$ running time. We include a proof in our supplement, but the outline involves breaking the process down into its smaller components, analyzing the time to aggregate the history, generate the rule neuron, perform the entailment calculation, and set the output variables.

\subsection{E. Training and Generation Process}
We conclude our presentation of \method by outlining how it works with the underlying generative architecture. \method is designed to integrate seamlessly into any underlying sequentially generating architecture. During both training and generation, the process involves feeding the patient record input into the underlying model to predict the probabilities of the variables in the next visit before updating those probabilities via the \method module.

During training, the true binary labels for the predicted visits are used as inputs to \method. The antecedent evaluation process updates the model's predicted probabilities based on logical constraints. The enhanced model's predictions, now incorporating constraint-driven logic, can then be compared to the ground truth labels, allowing for the calculation of a loss value that guides the model's training process without having to worry about learning prior 

During generation, the underlying model predicts the probabilities for the next visit based on the current record, and a full, binary visit is sampled from those probabilities. The predicted visit is then passed through \method, which enforces the logical constraints. This ensures that the generated sequence adheres to the desired constraints while still being synthesized by the underlying architecture.

So, \method enhances the underlying generative architecture by incorporating logical constraints seamlessly into both the training and generation processes. This enables the end-to-end training of a stronger model and the generation of logically constrained sequential data that maintains clinical coherence while meeting specific requirements.

\begin{figure*}[ht]
\centering
\includegraphics[width=0.87\textwidth]{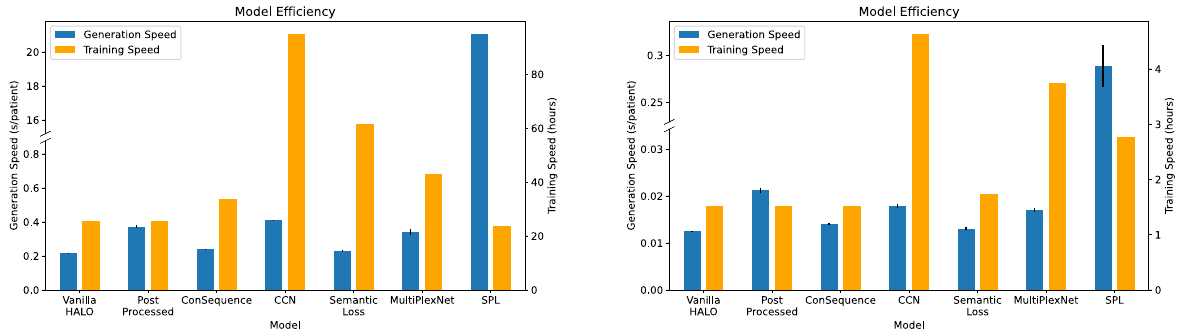}
\caption{Training and Generation Speeds on Outpatient (left) and Inpatient (right) datasets}
\label{fig:speeds}
\end{figure*}%

\section{Experiments}

\subsection{Experimental Setting}
We evaluate \method against state-of-the-art constraint enforcement approaches by applying each to the HALO \cite{theodorou2023synthesize} architecture (as one exemplar for these model agnostic techniques) for synthetic EHR generation. We perform a series of experiments on a pair of underlying real-world EHR datasets. Below are details concerning our data, baselines, and experimental design with remaining information found in our supplement.

\subsubsection{Datasets}
We used an outpatient EHR dataset from real-world US claims data and an inpatient EHR dataset from the public MIMIC-III dataset. Both datasets represent patients as visits. We encoded gender and age categories into the patient matrix representation $\mathbf{P}$ using binary variables and added a ``label visit" with only this demographic information at the beginning of each set of visits. This enables us to condition subsequent medical codes by the patient's gender or age.
Each dataset contains visits with ICD-9 diagnosis codes that we aggregated to higher level codes based on their respective ontologies. For instance, we combined diagnosis codes based on initial letter and first two digits (e.g., A24, A24.1, and A24.25 are aggregated to A24). We also limited each patient to a max of 100 visits, which removed very few visits in practice. Table \ref{tab:DataStats} provides the statistics of the final datasets. Additional details are available in the supplement.

\subsubsection{Rules} 
Although we expect experts to provide most rules in practice, we generated them in a data-driven approach here. 

We identify {\bf static rules} as:
1) For each demographic category, we permit at most one variable to be set (or none if unknown). Thus, autoregressive rules were added to prevent setting subsequent variables if a previous one was already set.
2) We identify pairs of codes that exclusively co-occur. For each pair, we establish a rule mandating the presence of the first code whenever the second one is present.

We identify {\bf temporal rules} as follows:
1) We identify prevalent codes that never appear within a demographic category. For each demographic category, we establish a rule disallowing these codes if the corresponding demographic variable is set in the first visit.
2) We detect common codes that only occur following a specific code. In each instance, we impose a rule preventing the former code unless the latter has previously occurred.
3) We find frequent codes that consistently appear in subsequent visits once they've occurred. For each case, we formulate a rule necessitating the code in the current visit if it was present in the previous one.

Our final rule sets consist of 71 rules (14 static, 57 temporal) for the outpatient dataset and 95 (11 static, 84 temporal) for the inpatient dataset. More details about cutoffs, counts by type, and some examples can be found in the appendix.

\subsubsection{Baselines}
We compare our method with several baseline approaches for handling logical constraints.
\begin{itemize}
    \item\textbf{Post Processing} is a naive approach removing any generated samples violating the constraints and generates more samples to replace them. It guarantees consistency with the constraints but is slower due to repeated generation and rule-checking, and may incur distribution shifts.    
    \item\textbf{Semantic Loss} \cite{xu2018semantic} is a regularization method based on the probability of breaking a constraint. It can handle temporal rules but does not guarantee compliance.
    \item\textbf{CCN} \cite{giunchiglia2021multi} is a rule-based method that uses maximum and minimum operations to set the output based on the values of the inputs in the antecedent. It guarantees compliance but cannot handle temporal rules or any where the output variable is set to 0.
    \item\textbf{MultiPlexNet} \cite{hoernle2022multiplexnet} introduces a categorical latent variable to select between satisfying assignments of constraints. It guarantees the satisfaction of static rules but does not support temporal constraints.
    \item\textbf{Semantic Probability Layer (SPL)} \cite{ahmed2022semantic} constructs probabilistic circuits to handle a wide variety of constraints. It can handle any static but no temporal rules\footnote{We modified both the SPL and corresponding HALO model for partial compatibility with our setting but couldn't achieve full capabilities. Further details are available in the supplement.}.
\end{itemize}

\subsubsection{Implementation}
We train HALO models on both datasets and apply each constraint enforcement method in turn. We generate a synthetic dataset of 10,000 patients from scratch with each method and without any constraints for comparison. These generations are performed on identical hardware (a single Tesla P100 GPU), and the generation process is repeated 25 times for each method (except for SPL on the outpatient dataset as it takes too long) to validate our results.


\subsection{Results on the Evaluation of Logical Adherence}
Our first evaluation measures each method's effectiveness in preventing rule violations. We count rule violations within the synthetic datasets, stratified by rule type, and calculate the percentage of fully valid generated patients.

Table \ref{tab:Violations} presents these statistics, highlighting that \method is the only non-post-processing method guaranteeing full prevention of rule violations. Although CCN, MultiPlexNet, and SPL adhere to some static rules and Semantic Loss regularizes all constraints, no baseline methods fully satisfy temporal rules, leading to numerous invalid generations. In contrast, \method manages all rule types, preventing invalid generations more efficiently than the naive Post Processing approach.

\subsection{Results on the Evaluation of Data Enrichment} 

This section assesses whether knowledge infusion impacts learning and generation by evaluating the quality of the model and its synthetic data.
We first calculate the model's perplexity on a held-out test set, which signifies the log probability of the test set, normalized by the number of codes in the dataset. Lower perplexity values indicate superior performance. Table \ref{tab:Perplexities} displays the perplexities for each method on each dataset. We find that \method enhances the model's performance on both datasets by eliminating error sources from predicted rule violations, allowing for absolute non-violation predictions, and facilitating the model to learn dataset patterns during training beyond the rules.
\method exhibits the most significant improvement, while CCN and Semantic Loss moderately enhance learning. In contrast, MultiPlexNet and SPL baselines hinder the modeling of the underlying data distribution by introducing more complex loss formulations during training.

\begin{table}
    \centering
    \small
    \begin{tabular}{lcc} \toprule
         & Outpatient  & Inpatient  \\\midrule
    Vanilla HALO 
    & 24.019     & 13.443 \\\hline
    Semantic Loss 
    & 23.995     & 13.432 \\
    CCN 
    & 23.976     & 12.985\\
    MultiPlexNet 
    & 119.560     & 24.260\\
    SPL 
    & 34.454     & 22.794\\\hline
    \method           & \textbf{23.924}   & \textbf{12.712}\\\bottomrule
    \end{tabular}
    \caption{Modeling Perplexity Performance}
    \label{tab:Perplexities}
\end{table}

We also evaluated the realism of the generated dataset by comparing the probabilities of codes and their combinations in each synthetic dataset to their original training datasets, and we provide those results in our supplement.

\subsection{Results on Method Efficiency} 
The final evaluation checks if \method maintains robust real-world efficiency.
Figure \ref{fig:speeds} presents the training and generation speeds for each model. Our results highlight that \method provides commendable real-world efficiency. It outpaces other constraint methods during training, while the baselines slow down due to memory or algorithmic inefficiencies. During generation, \method surpasses all baselines barring Semantic Loss, which matches Vanilla HALO model during inference but can't guarantee constraint adherence. \method experiences less than a 13\% slowdown in generation time on both datasets, showing that it meets the real-world efficiency criterion.

\section{Conclusion}
We introduce \method, an effective approach that enforces logical constraints in sequential generative models for the first time. \method seamlessly integrates rule-based domain knowledge into generative neural network outputs through temporal aggregation and antecedent evaluation modules. It efficiently enforces these constraints using a matrix multiplication formulation of these modules, ensuring the satisfaction of both hard and soft logical constraints. Through extensive experimentation in the domain of electronic health record generation, we demonstrate that \method outperforms comparable models in both efficiency and effectiveness, incurring a minimal slowdown compared to an unconstrained model while eliminating all rule violations and enhancing overall generative quality.

\section{Ethics Statement}
We view the possible broader impact of \method through the lens of enabling better and more realistic sequential synthetic data. This includes facilitating better video generation, more effective synthetic financial data, and improved healthcare data, These increases in generative quality alone can have transformative effects on machine learning research in related domains, improve the creative development of media, and permit freer data sharing. However, even beyond overall improvements in data quality, we see the elimination of logical inconsistencies as especially effective for increasing adoption of these generative models and the data they produce. One of the typical checks for utilizing such systems, particularly in the healthcare domain, is a manual review of the created data. Logical inconsistencies are particularly visible and therefore among the main barriers in the adoption process. Therefore, we believe that \method's ability to prevent such inconsistencies can facilitate not just better generative models but allow them to disseminate more widely and exponentially propagate the positive impact of synthetic data.

\bibliographystyle{aaai24}
\bibliography{refs}


\end{document}


\maketitle

Here we provide additional experimental notes, details, and results beyond that which we could fit into our main paper. We hope that this provides complete transparency and reproducibility of our method, experiments, and results.

\section{Method}
\subsection{Full Set of Notations}
We provide a full table of notations for reference in Table \ref{fig:Notation}.

\subsection{Runtime Analysis}
\method can process, evaluate, and adhere to rules in $O(T^2|C|)$ running time. We include the full proof here.

\begin{proof}
Given that we are in the temporal generation stage, we cannot use a matrix formulation since we must correct our outputs at every step. Assuming that applying $f_{I^{(r)}}$ takes $O(T)$ time, forming $m$ takes $O(T)$ time. The formation of the history vector requires calculation of the logical or of $O(T)$ vectors of size $|C|$, and thus is $O(T|C|)$. Constructing the rule neuron can be done in $O(|C|)$ time, assuming constant time lookups, which can reasonably be implemented for a small rule size. The activation process requires the element-wise multiplication of two vectors of size $|C|$, and thus should take $O(|C|)$ time. Ultimately, the runtime per generation step should be $O(T|C|)$.

For the training stage, we are able to employ the matrix formulation. The formation of $\mathbf{M}^{(r)}$ would require $O(T^2)$ time at minimum, given that $\mathbf{M}^{(r)} \in \mathbb{R}^{T \times T}$. Computing $\mathbf{M}^{(r)}\mathbf{P}$, the product of a matrix within $\mathbb{R}^{T\times T}$ and a matrix within $\mathbb{R}^{T \times |C|}$, would naively take $O(T^2|C|)$ time. Constructing $\mathbf{W}^{(r)}$ should take $O(|C|)$ time, given the sparsity of the matrix. Due to this sparsity, the matrix product $\mathbf{X}^{(r)}\mathbf{W}^{(r)}$ can also be naively calculated in $O(T|C|)$ time, similarly to a matrix vector product. Checking equality of two vectors in $\mathbb{R}^T$ and setting values can be done in $O(T)$ time. Thus the runtime for the whole process would be $O(T^2|C|)$. However, assuming parallelization and a more 
efficient matrix multiplication module, this runtime can be significantly reduced.
\end{proof}

\begin{table}
\smaller
\centering
\begin{tabular}{c|p{0.7\columnwidth}} \toprule
Notation   & Description \\ \midrule
$\mathcal{P}$          & A patient's health record sequence \\
$\mathcal{V}^{(t)}$          & A patient's $t$-th visit in their record \\
$c^{(t)}_i$          & The $i$-th 
medical code in the $t$-th visit \\
$T$                 & The total number of visits in $\mathcal{P}$ \\
$\mathcal{C}$             & The set of all medical codes \\\hline
$\mathbf{P} \in \mathbb{R}^{T \times |C|}$          & The matrix representation of $\mathcal{P}$ \\
$\mathbf{v}_t \in \mathbb{R}^{|C|}$          & The vector representation of the $t$-th visit \\
$c_t^i$             & Binary variable for the presence or absence of the $i$-th code in $\mathcal{C}$ in $\mathbf{v}_t$ \\\hline
$\mathcal{P}'$          & A synthetic patient record\\
$\mathcal{D}$           & The dataset provided to train the model to learn $P(\mathbf{P})$\\
$\mathcal{R}$           & The set of logical constraints over the possible code combinations. \\\bottomrule
\end{tabular}
\caption{Notations used in \method}
\label{fig:Notation}
\end{table}

\section{Experimental Details}
\subsection{Source Code}
We release the code for our model and inpatient experiments at \url{https://github.com/btheodorou/KnowledgeInfusion}.

\subsection{Dataset Construction}
While the outpatient EHR claims dataset is proprietary belonging to a large clinical trial company and so not publicly available, the inpatient dataset is the public MIMIC-III ICU stay dataset \cite{MIMIC} and can be accessed on PhysioNet after a required training and application. A script for processing the raw dataset to be used within our experiments is provided in our source code.

\begin{table*}[]
    \centering
    \smaller
    \begin{tabular}{lcccc}
    \toprule
    & \multicolumn{2}{c}{Outpatient Dataset} & \multicolumn{2}{c}{Inpatient Dataset} \\
    & Training  & Generation (s/patient) & Training & Generation  (s/patient)         \\ \midrule
    Vanilla HALO         & 25:26:31       & 0.218 $\pm$ 0.001       & 1:30:38       & 0.0124 $\pm$ 0.0001  \\\hline
    Post Processing         & 25:26:31       & 0.371 $\pm$ 0.008        & 1:30:38       & 0.0212 $\pm$ 0.0004  \\
    Semantic Loss     & 61:28:37       & 0.229 $\pm$ 0.004       & 1:44:16       & 0.0130 $\pm$ 0.0002  \\
    CCN              & 94:59:22       & 0.411 $\pm$ 0.001       & 4:38:13       & 0.0179 $\pm$ 0.0004  \\
    MultiPlexNet      & 35:09:16       & 0.340 $\pm$ 0.004        & 3:44:53       & 0.0170 $\pm$ 0.0004  \\
    SPL      & 23:39:00       & 21.378        & 2:46:09       & 0.2689 $\pm$ 0.0045  \\\hline
    \method           & 33:40:28   & 0.238 $\pm$ 0.001   & 1:30:50   & 0.0140 $\pm$ 0.0002 \\\bottomrule
    \end{tabular}
    \caption{Wall Clock Benchmarking}
    \label{tab:Generation}
\end{table*}

\subsection{Rule Generation}
As mentioned in the main paper, we used five different data-driven approaches to generating rules. Here we provide the prevalence cut-offs used, rule counts, and rule examples found by rule type. Note that the cut-offs are set on a per-rule basis to balance the desire to find rules with the desire to prevent an explosion of noisy rules. Additionally, the thresholds in the outpatient dataset are universally greater than or equal to those in the inpatient dataset due to the greater number of patients there.\\
\begin{itemize}
    \item{Static Rules \begin{itemize}
        \item We required phecodes that only appeared simultaneously with other codes to occur 10 times overall in both the inpatient dataset and outpatient datasets. We found 4 and 3 such rules for the inpatient and outpatient dataset respectively. These rules more than any other category appeared to be correlational rather than clinical mandates upon examination (for example, certain conditions only occurring in the outpatient dataset if the patient was impoverished).
        \item We had 7 autoregressive rules limiting patients to at most one demographic variable from each category in the inpatient dataset and 11 such rules in the outpatient dataset. These were acquired as part of our representational format. For example, patients can not have both the male and female variables nor be born in both the 1980-1999 and 2000-2020 buckets.\\
    \end{itemize}}
    \item{Temporal Rules \begin{itemize}
        \item We required phecodes that never appear for given demographic groups to occur 500 times overall in both the inpatient and outpatient dataset. We also add any additional codes even if they are below that 500 appearance threshold if they can ``force'' a code that is above that threshold to appear via the first type of static rule. We found 83 and 57 such rules for the inpatient and outpatient dataset respectively. For example, we found that women can not have prostate cancer.
        \item We required phecodes that never appear without a separate phecode occurring previously to occur 10 times overall in the inpatient dataset and 100 times in the outpatient dataset. We found no such rules for either of the inpatient and outpatient datasets.
        \item We required phecodes that once they occur always continue to occur to show up 10 times overall and 5 times repeated in the inpatient dataset and 500 times overall with 250 times repeated in the outpatient dataset. We found 1 and 0 such rules for the inpatient and outpatient dataset respectively. The found rule was that patients with spina bifida always maintain the condition in future visits (a finding in line with the medical reality of spina bifida as a lifelong condition).
    \end{itemize}}
\end{itemize}

\subsection{HALO}
While \method and the compared methods are model (and even task) agnostic, we use the HALO \cite{theodorou2023synthesize} architecture as an example model with which to test and compare the performance of our constraint enforcement module. However, we note that the experiments could be replicated with any sequentially generative architecture such as \cite{SynTEG}, RNNs, or any number of general predictive models adapted to become generative models to similar effect.

However, given we use HALO throughout our paper, we provide a brief overview of the model here. HALO is a language model-style architecture for EHR generation which uses a stack of transformer decoder blocks to aggregate historical visits into a patient history representation before using that representation in conjunction with the current visit to perform more granular autoregressive modeling at the code-by-code level via a stack of masked linear layers. Given this code-level modeling, HALO generates visits and then records by looping through one code at a time. However, to avoid this intricacy and make it more representative of sequentially generating methods in general (which synthesize one visit at a time for EHRs or one ``token'' at a time for other applications such as text generation), we group the code-level generations into visit generation steps and apply the constraint enforcement modules after this larger step.

\begin{table*}[]
    \centering
    \smaller
    \begin{tabular}{lcccccc}
    \toprule
    & \multicolumn{3}{c}{Code Correlation on Outpatient Dataset} & \multicolumn{3}{c}{Code Correlation on Inpatient Dataset} \\
    & Individual  & Co-Occurring  & Sequential  & Individual  & Co-Occurring  & Sequential        \\ \midrule
    Vanilla HALO   & 0.940 $\pm$ 0.0014 & 0.866 $\pm$ 0.0051 & 0.904 $\pm$ 0.0026
    & 0.964 $\pm$ 0.0009 & 0.929 $\pm$ 0.0015 & 0.911 $\pm$ 0.0030    \\\hline
    Post Processing             & 0.932 $\pm$ 0.0017 & 0.835 $\pm$ 0.0067 & 0.892 $\pm$ 0.0030
       & 0.886 $\pm$ 0.0031 & 0.878 $\pm$ 0.0040 & 0.934 $\pm$ 0.0024 \\
    Semantic Loss     & 0.919 $\pm$ 0.0017 & 0.820 $\pm$ 0.0059 & 0.874 $\pm$ 0.0032
      & 0.967 $\pm$ 0.0012 & 0.955 $\pm$ 0.0015 & 0.929 $\pm$ 0.0031    \\
    CCN             & \textbf{0.967 $\pm$ 0.0011} & \textbf{0.943 $\pm$ 0.0028} & \textbf{0.947 $\pm$ 0.0020}
         & \textbf{0.993 $\pm$ 0.0003} & \textbf{0.980 $\pm$ 0.0005} & \textbf{0.941 $\pm$ 0.0019}    \\
    MultiPlexNet     & 0.099 $\pm$ 0.0035 & -1.89 $\pm$ 0.0171 & -0.082 $\pm$ 0.0062
        & 0.799 $\pm$ 0.0021 & 0.763 $\pm$ 0.0024 & 0.778 $\pm$ 0.0034    \\
    SPL              & 0.738     & 0.468     & 0.407     & 0.988 $\pm$ 0.0002     & 0.838 $\pm$ 0.0013     & 0.880 $\pm$ 0.0026 \\\hline
    \method           & 0.956 $\pm$ 0.0013 & 0.934 $\pm$ 0.0036 & 0.914 $\pm$ 0.0030
         & 0.975 $\pm$ 0.0004 & 0.948 $\pm$ 0.0006 & 0.920 $\pm$ 0.0018     \\\bottomrule
    \end{tabular}
    \caption{Phecode Correlation $R^2$ Scores}
    \label{tab:correlation}
\end{table*}

\subsection{Hyperparameters}
We copy most of the hyperparameters (model embedding and layer size, initialization ranges, learning rate, epochs, etc.) from the original HALO source code released by the original authors of \cite{theodorou2023synthesize}. However, we make two key adjustments in adding a patience hyperparameter for early stopping and then adjusting the batch size to the largest value which could fit in memory for our CCN baseline (which is memory inefficient, although we still aimed to have each compared method use the same batch size for a fair efficiency and performance comparison). All hyperparameters can be found in our source code.

\subsection{Compute}
We utilize a single Tesla P100 GPU with 32 GB of RAM for all experiments. Using this, the GPU-hours needed to conduct our experiments can be directly calculated from our results as we directly measure wall clock time. Specifically, combining the training time and generation time (multiplying by the 10,000 generated records and the 25 separate generation runs for all model-dataset pairs except the SPL baseline on the outpatient dataset) covers the vast majority of the time spent. Additional analyses (number of violations, test set perplexity, code correlation statistics) either do not require a GPU or take extremely minimal time (less than an hour total) with one. Using this approach, we estimate that 504 GPU-hours were spent conducting our experiments.

\subsection{Adapting the Semantic Probabilistic Layer}
The SPL baseline \cite{ahmed2022semantic} is one of the most exciting recent works on logically constrained models. However, it is incompatible with the full HALO model due to its reliance on intermediate embeddings as inputs to its probabilistic layers rather than whole model outputs. If we were to use the full HALO output as the ``embedding," it would result in data leakage caused by HALO's code-level autoregressive modeling. To address this, we use the previous embedding, but we note that this approach weakens the underlying model compared to other baselines.

\section{Results}

\subsection{Error Calculations}
While repeating the full experiments with separately trained models was too computationally inefficient to perform, we did generate multiple datasets with different random seeds to provide a measure of variance and confidence for our main results. Specifically, for each compared model on each dataset, we generated 25 different datasets of 10,000 synthetic patient records. We used those 25 different runs to provide different estimates of generation speed, violation prevention, and dataset quality. For each corresponding metric, we report the mean value over those runs as well as a 95\% confidence interval calculated as 1.96 standard errors (standard deviation divided by the square root of the number of runs) from that mean. Note that we do not provide any error bounds for the SPL baseline results on the outpatient dataset as it was computationally prohibitive to perform more runs.

\subsection{Full Generation Speeds}
We provided bar plots of training and generation speeds in our main paper as we felt that it was an easier medium to visualize and compare the results across each method at once. We provide the full results here in Table \ref{tab:Generation} for completeness.

\subsection{Code Plots}
\begin{figure*}
    \centering
    \includegraphics[scale=0.65]{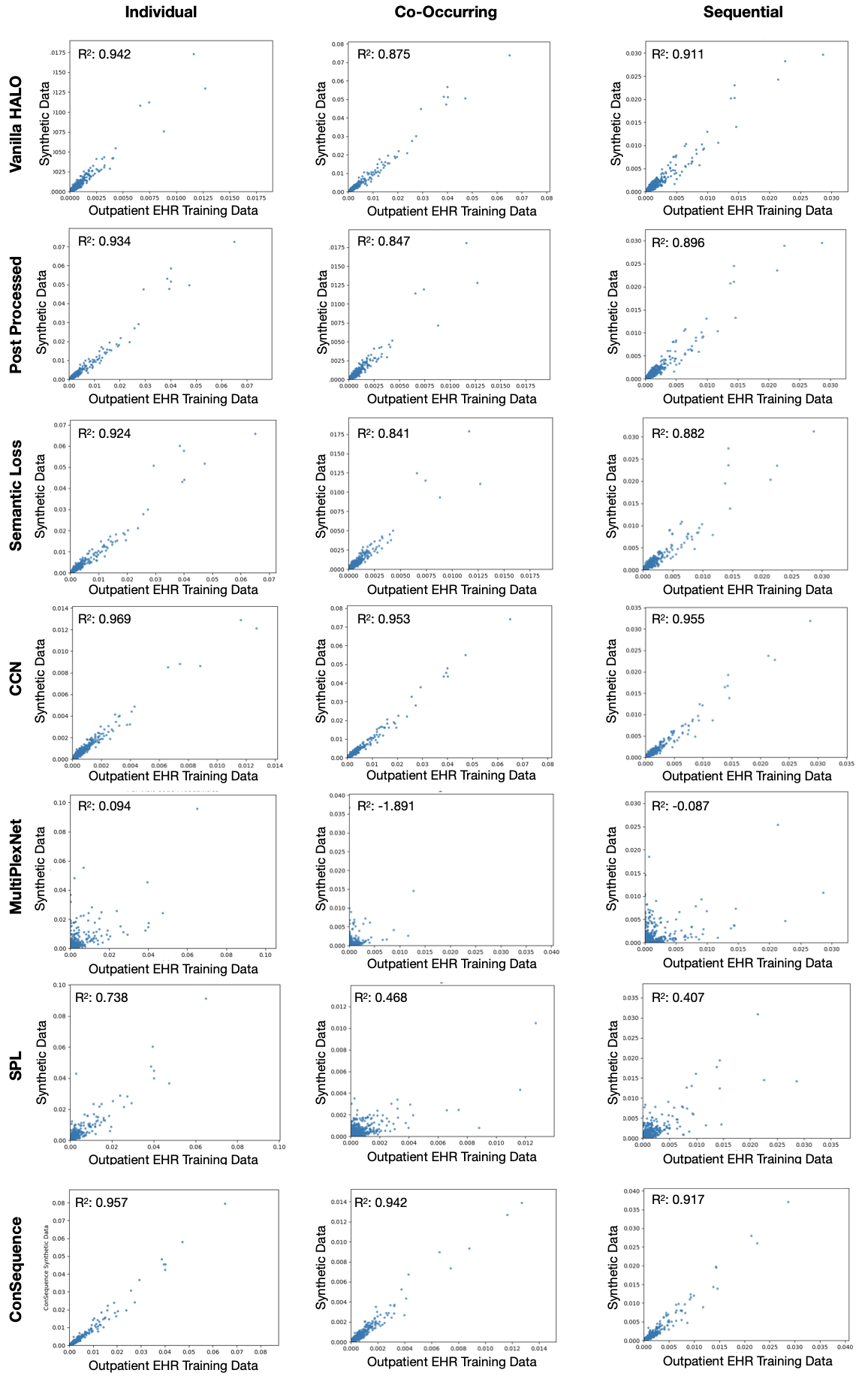}
    \caption{Code correlation plots for each compared method on the outpatient dataset}
    \label{fig:Probabilities}
\end{figure*}

Our core evaluation of model quality in our main paper was perplexity. However, we also evaluated the realism of the generated datasets by comparing the probabilities of individual codes, code co-occurrences, and sequentially paired codes in each synthetic dataset with those within their original training datasets. We display the $R^2$ scores of the correlation between these probabilities in Table \ref{tab:correlation}, and we provide an example of the full probability plots for the outpatient dataset for each compared method in Figure \ref{fig:Probabilities}.
These results largely mirror those from the perplexity evaluation. \method performs consistently better than the unconstrained HALO model in generative performance while the Semantic Loss baseline also offers improvement in the inpatient (though not outpatient) setting. Here, however, CCN achieves the best performance in offering even larger gains in generative performance than \method. Additionally, we can see here more clearly the detrimental effect that post-processing (the only other approach besides \method capable of preventing all rule violations) has on generative quality, causing a distribution shift that lowers the code correlations. Finally, MultiPlexNet and SPL's worse perplexities are matched by poor correlation statistics, especially in the more complex outpatient dataset. Overall, we see that \method not only maintains but improves generative performance through its constrained training approach. This, along with it being the only method to prevent all rule violations and doing so with excellent efficiency, makes it the best of all the compared methods.

\bibliographystyle{aaai24}
\bibliography{refs}
